\documentclass[%
 aip,
 amsmath,amssymb,
 reprint,%
 floatfix,
]{revtex4-1}

\usepackage{graphicx}
\usepackage{dcolumn}
\usepackage{bm}
\usepackage{color}
\usepackage{newtxtext,newtxmath}
\usepackage[utf8]{inputenc}
\usepackage[T1]{fontenc}
\usepackage{etoolbox}
\usepackage{ulem}

\usepackage{amsmath}

\makeatletter
\def\@email#1#2{%
 \endgroup
 \patchcmd{\titleblock@produce}
  {\frontmatter@RRAPformat}
  {\frontmatter@RRAPformat{\produce@RRAP{*#1\href{mailto:#2}{#2}}}\frontmatter@RRAPformat}
  {}{}
}%
\makeatother
\begin{document}

\preprint{AIP/123-QED}

\title{Data Driven Modeling for Self-Similar Dynamics}
\author{Ruyi Tao}
\email{taoruyi@mail.bnu.edu.cn}
\affiliation{ 
School of Systems Science, Beijing Normal University
}%
\affiliation{Swarma Research}%
\author{Ningning Tao}%
\affiliation{ 
School of Systems Science, Beijing Normal University
}%

\author{Yi-zhuang You}
\affiliation{%
Department of Physics, University of California San Dieg
}%

\author{Jiang Zhang}
\email{zhangjiang@bnu.edu.cn}
\affiliation{ 
School of Systems Science, Beijing Normal University
}%
\affiliation{Swarma Research}%


\begin{abstract}
Multiscale modeling of complex systems is crucial for understanding their intricacies. Data-driven multiscale modeling has emerged as a promising approach to tackle challenges associated with complex systems in recent years. But at present, this field is more focused on the prediction or control problems in specific fields, and there is no suitable framework to help us promote the establishment of complex system modeling theory. On the other hand, self-similarity is prevalent in complex systems, not only revealing interesting properties of complex systems, but also hinting that large-scale complex systems can be modeled at a reduced cost. In this paper, we introduce a multiscale neural network framework that incorporates self-similarity as prior knowledge, facilitating the modeling of self-similar dynamical systems.  For deterministic dynamics, our framework can discern whether the dynamics are self-similar. For systems with stochastic dynamics, our framework is capable of comparing and identifying which parameter set approximates self-similarity more closely. The case in critical dynamical system also demonstrates that our framework can identify critical regions. Additionally, it facilitates the extraction of scale-invariant kernels from the dynamics, enabling modeling at any scale based on the assumption of homogeneity. Importantly, the incorporation of self-similarity priors empowers our model to autonomously learn the coarse-graining strategy. This ability is instrumental in identifying power-law exponents within self-similar systems, offering significant contributions to the development of both theoretical and practical approaches in complex system modeling.
%
\end{abstract}

\maketitle

\begin{quotation}


Modeling complex systems is crucial for understanding, predicting, and even controlling them. What sets complex systems apart from other systems is the notion that simple mechanisms at the micro level can lead to complex properties at the macro level. Thus, a key to understanding these phenomena lies in observing complex systems from multiple scales, making multiscale modeling essential for capturing the fundamental properties of these systems. On the other hand, many complex systems exhibit self-similarity, indicating that, in many cases, they are scale-free. This paper aims to integrate self-similarity as prior knowledge and uses neural network techniques for multiscale modeling of complex systems. One of the advantages of this approach is its ability to identify whether a system is self-similar. Our primary focus is on dynamical systems. Through a learnable approach, we can gauge whether a dynamical system exhibits self-similar dynamics based on the final error assessment. Our framework is applicable to both deterministic and stochastic systems. Moreover, with our framework, we can identify a scale-invariant dynamical kernel. This means we not only discover a scale-independent feature of a dynamical system, but can also use the learned kernel to simulate the system at any scale, greatly reducing computational costs.

\end{quotation}

\section{\label{sec:level1}Introduction}

Complex systems modeling is essential for understanding, predicting, and even controlling a complex system. Due to the non-linear, self-organizing, emergence, and other complex behaviors in them, modeling complex systems has always been challenging. In recent decades, data-driven approaches, leading by machine learning, have shown significant advantages in so many fields, which inspired us to do better in modeling complex systems. On the other hand, self-similarity is a common feature of complex systems. From natural systems, like the fractal structure of vegetation clusters in the Amazon rainforest and the Tibetan plateau \cite{taubert2018global}, the critical phenomena in atmospheric precipitation \cite{Peters2006}, to societal systems like network traffic \cite{smith2011dynamics}, the avalanche of public opinion in social medias \cite{Notarmuzi2021}, and neural system like critical phenomenon in brain \cite{Herculano-Houzel2014,LaRocca2018,Ribeiro2021,Grosu2023} and so on, there are so many evidences of scale-invariant properties in complex systems. Thus, we're motivated to integrate self-similarity as prior knowledge, aiming for data-driven multi-scale modeling of complex systems.

Two aspects of modeling complex systems are network structure and dynamics. The concept of self-similarity in networks was proposed in \cite{Song2005}, leading to explorations of multiscale network modeling \cite{Radicchi2008,Radicchi2009,Rozenfeld2010,Garcia-Perez2018,Chen2021,Chen2023,Villegas2023}. This offers insights into modeling large-scale systems more cost-effectively. Discussions integrating dynamics with self-similarity in complex system modeling have been relatively scarce. By combining multiscale dynamics with self-similarity, maybe we can model complex dynamic systems with fewer parameters, and capture the scale-invariant key features.


A related field is Reduced-Order Model (ROM)\cite{Antoulas2004,LUCIA200451,Luchtenburg2021,schmid2022dynamic,brunton2021modern}, which seeks lower-dimensional representations of systems and their dynamics. It aims to reproduce the linear\cite{} or non-linear\cite{} high-dimensional dynamics at microscopic level with simpler models according to capture essential features. However, this kind of method is often used more in engineering prediction and simulation, and usually does not pay attention to the physical meaning of the reduced order model.

A similar but more common method in physics is the Equation-Free Method (EFM) from Kevrekidis\cite {kevrekidis2003equation, kevrekidis2004equation,williams2015data}, which indeed care about physical meaning of macrostate after coarse-grained. This approach aims to bypass direct mathematical descriptions of complex systems, using microsimulations to understand and predict macroscopic behavior through interactions between macroscopic and microscopic scales. Although not explicitly stated, it also implies the dynamics has self-similarity of the predicted results at different scales. But they didn't discussed the concept of self-similarity. 




Another related topic is renormalization group (RG) theory \cite{kadanoff1966scaling,wilson1983renormalization,pelissetto2002critical}. RG theory deals with how system parameters vary with scale, which is essential in condensed matter physics for addressing phase transitions and critical phenomena, and it does touch upon self-similarity. In this context, self-similarity means the system remains unchanged at a fixed point in parameter space. RG theory has been successful for solving the problems in equilibrium systems. Furthermore, Theoretical physicists also invented the theory of dynamic RG to describe how a dynamic system parameters change with scale \cite{schollwock2005density,Tauber2014,Cavagna2017}, and try to use it to solve the critical problems in non-equilibrium systems. However, due to the complexity of dynamic systems, the analysis of dynamic RG involves intricate mathematical techniques. On one hand, it heavily depends on the researcher's expertise. On the other, it's currently limited to analyzing very basic systems, like kinetic Ising model \cite{Yalabik1982,Mazenko1981}, sandpile model \cite{Storer1995,Ivashkevich1999,Lin2002} and Viseck model \cite{Cavagna2019a,Cavagna2019,Cavagna2021},etc\cite{moise1999renormalization,Israeli2006}. And There are also subjective renormalization strategies, where different strategies might lead to varying results. It also adds difficulty to the theory's direct application in complex dynamic systems.


Hence, analytical solutions and simple models are insufficient for complex systems, but data-driven methods might be the way forward. Physics-informed machine learning has seen recent success in numerous domains \cite{Karniadakis2021,Wright2022,Vinuesa2022,raissi2019physics}. Designing neural network architectures based on inductive biases is considered the most principled way of making a learning algorithm informed by physics \cite{Karniadakis2021}. Among them, there are many multi-scale neural network architectures are introduced for improved predictions and simulation inspired by the ROM\cite{lusch2018deep,azencot2020forecasting,williams2015data,bevanda2021koopman,otto2021koopman,brunton2020machine,vlachas2018data} and EFM\cite{kevrekidis2004equation,Vlachas2022,williams2015data,feng2023learning}. But these frameworks focus on specific tasks and don't discuss the system's self-similarity features. At the same time, integrating renormalization theory and machine learning has seen progress \cite{Walker2020,Shiina2021,Hu2020,Hou2023,Ron2021,Chung2021,Li2018,Lenggenhager2020,Iso2018}, but most research focuses on static systems without considering dynamics. 

Based on the above background, we introduce a multi-scale neural network framework that integrates self-similarity priors for modeling complex dynamic systems.

Our contributions include:
\begin{itemize}
\item By introducing self-similarity as the prior information of model design, our framework can identify if the dynamic systems are self-similar. For deterministic dynamics, our framework can discern whether the dynamics are self-similar. For stochastic dynamics, it can compare and determine which parameter set is closer to self-similarity

\item If the dynamics is self-similar, our framework can automatically learn the dynamic renormalization strategies near the parameter of fix point.


\end{itemize}

The structure of the article is as follows: In section \ref{se:method}, we give the formal definition of self-similar dynamics, and give our model framework design idea. In the experimental part, we use two examples to illustrate the feasibility of the model in section \ref{se:experiments}. At last, we discuss the potential of our framework and its implications for understanding complex dynamical systems.

\section{Method}
\label{se:method}
\subsection{Definition}
\label{se:definition}

Self-similar dynamics, intuitively, means the system's dynamics are consistent or similar across multiple scales. This doesn't just refer to the form of the dynamics but also their parameters. Based on this intuition, let's provide a formal definition. We define the microstates of a system as $\boldsymbol{X} \in \mathbb{R}^{n^d}$, where $n^d$ represents the size of the space $\boldsymbol{X}$ occupies, indicating the possible $n^d$ components of $\boldsymbol{x_i}$, and $d$ is the spatial dimension of the system. Its micro dynamics is given by $\Delta \boldsymbol{X} = f(\boldsymbol{X},\boldsymbol{W})$, where $\boldsymbol{W}$ is the parameters of the dynamics. In our discussion, self-similarity dynamics means that after a system evolves over time following states $\boldsymbol{X}_{0:t_m} = \{\boldsymbol{X}_0,\boldsymbol{X}_{1},\dots,\boldsymbol{X}_{t_m}\}$, and is transformed by mapping $P$ to produce macrostate $\boldsymbol{Y} \in \mathbb{R}^{N^d \times F}$ such that $\boldsymbol{Y} = P ({\boldsymbol{X}_{0:t_m}})$. Here, $N^d$ represents the size of the space $\boldsymbol{Y}$ occupies, indicating the possible $N^d$ components of $\boldsymbol{y_i}$. In general, $N<n$. By doing so, in theory, there should also exist a dynamical equation $\Delta \boldsymbol{Y} = f'(\boldsymbol{Y},\boldsymbol{W'})$ describing the evolution of the macroscopic variable $\boldsymbol{Y}$. The point is both the form and parameter of the micro dynamics remain unchanged or similar to the one in the macro dynamics. That is $f' = f$ and $\boldsymbol{W}' = \boldsymbol{W}$.  

More strictly, for every microstate component $\boldsymbol{x^i}$, the dynamics satisfies

\begin{equation}
\label{eq:dynamic_assumption}
{\Delta \boldsymbol{x^i}} = f^i(\boldsymbol{x^i},\{\boldsymbol{x^j},\forall \boldsymbol{x^j} \in Nei(\boldsymbol{x^i})\},\boldsymbol{w^i})
\end{equation}

Here,$f^i$ represents the dynamics of component $\boldsymbol{x^i}$. The set $\{\boldsymbol{x^j},\forall \boldsymbol{x^j} \in Nei(\boldsymbol{x^i})\}$ denotes the neighboring states of $i$. $\boldsymbol{w^i}$ is the dynamic parameter for each component $i$. Our discussion implicitly assumes that the dynamics are homogeneous, that is all components share the same dynamic parameter $\boldsymbol{w}$. Please note that the $\boldsymbol{w}$ may not necessarily be local. If the interactions take place within a network in non-Euclidene space, then $\boldsymbol{w}$ represents the interaction parameter among neighboring elements in that space, potentially encompassing higher-order interactions.

The mapping $P$ is a neighborhood operation that preserves component invariance, which means all components could share the same mapping operator. Here we denote $P^i$ to represent coarse-graining mapping for each component $i$.  Let's take Markov process in euclidean space as example to describe this coarse-graining process.(show as Fig \ref{fig:framework_cnn}(a)). Let $\boldsymbol{\Tilde{X}}^i_t \in \mathbb{R}^{S^d \times T} = \{\boldsymbol{x}^{i-S^d}_{t-T},\boldsymbol{x}^{i-S^d+1}_{t-T},\dots,\boldsymbol{x}^{i}_{t-T+1},\dots,\boldsymbol{x}^{i}_{t}\}$ be a basic unit within $\boldsymbol{X}_{0:t_m}$. It captures the local system state over spatial scale $S$ and temporal scale $T$. So $\boldsymbol{X}_{0:t_m}$ can be represented as a collection of these unites, that is $\boldsymbol{X}_{0:t_m} = \{\boldsymbol{\Tilde{X}}^0_0,\boldsymbol{\Tilde{X}}^1_0,\dots,\boldsymbol{\Tilde{X}}^{n^d/S^d}_0,\boldsymbol{\Tilde{X}^0_1},\dots,\boldsymbol{\Tilde{X}}^{n^d/S^d}_{t_m\//T} \} $. For each $\boldsymbol{\Tilde{X}}^i_t$ we have

\begin{equation}
\label{eq:coarse_grain_assumption}
\boldsymbol{y^{i}_{t}}= P^{i}(\boldsymbol{\Tilde{X}}^i_t), P^i \in \mathbb{R}^{S^d \times T \times 1}
\end{equation}

The operator $P^i$ maps from $S^d \times T$ to 1, as illustrated in Fig.\ref{fig:framework_cnn}(a). Therefore, the global mapping result is equivalent to concatenating each local mapping outcome, which means

\begin{equation}
\label{eq:coarse_grain_cat}
P(\boldsymbol{X}) = P^0(\boldsymbol{\Tilde{X}}^0_0) \bigoplus P^1(\boldsymbol{\Tilde{X}}^1_0) \bigoplus \dots \bigoplus P^{n^d/S^d}(\boldsymbol{\Tilde{X}}^{n^d/S^d}_{t_m\//T} \} )
\end{equation}

\begin{figure}
    \centering
    \includegraphics[width=1\linewidth]{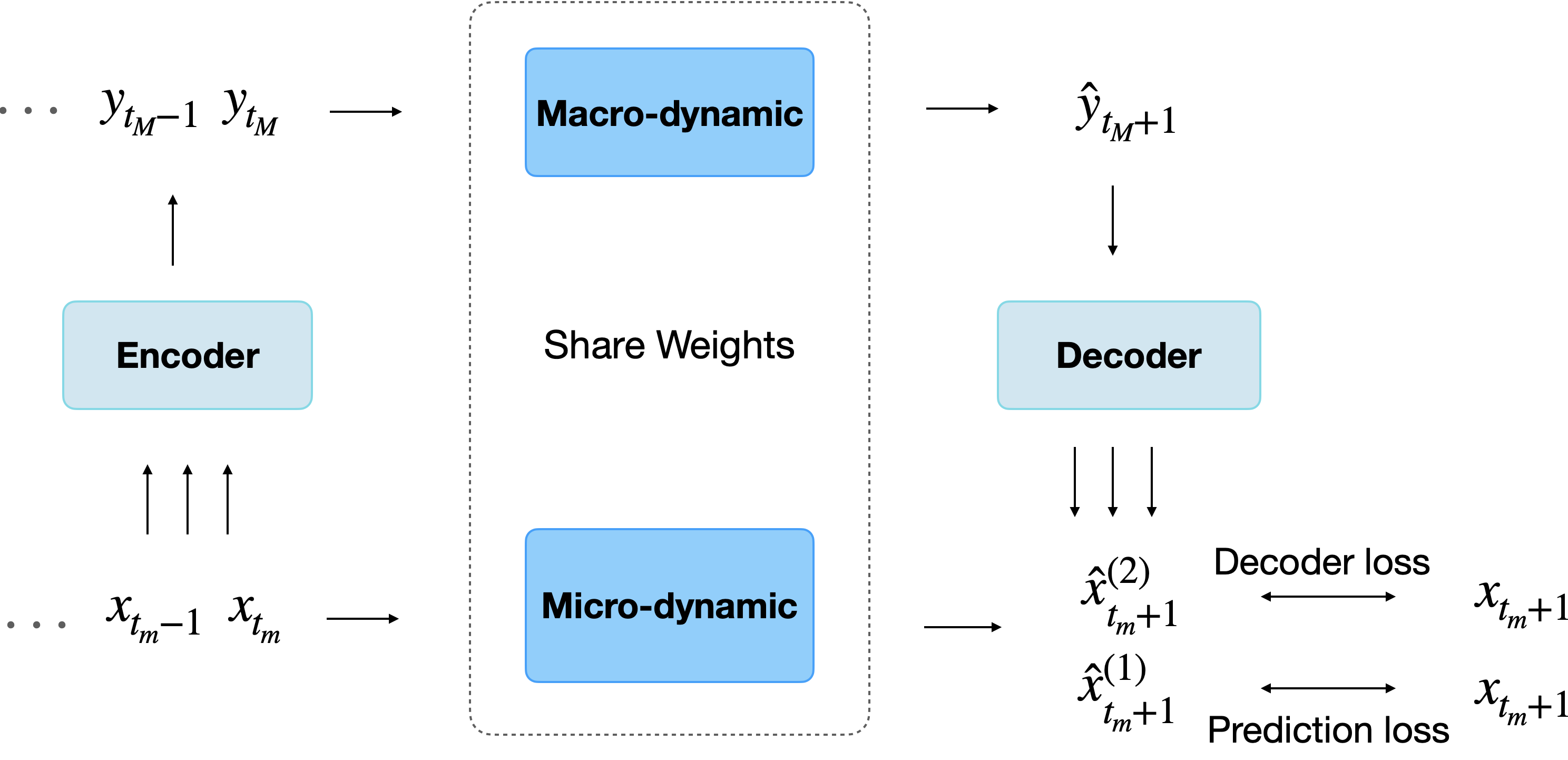}
    \caption{\textbf{Framework}}.
    \label{fig:framework}
\end{figure}

So in simpler terms, self-similar dynamics means that for every local state $\boldsymbol{\Tilde{X}}^i_t$ of the microscopic system, once mapped to the macrostate $\boldsymbol{y^i_t}$ by $P^i$, the form and parameters $\boldsymbol{w}^i$ of each macro-state dynamics $f^i$ remain consistent with the micro one. That is, the macro dynamics also follow $\boldsymbol{\Delta y^i} = f(\boldsymbol{y^i},\{\boldsymbol{y^j},\forall \boldsymbol{y^j} \in Nei(\boldsymbol{y^i})\},\boldsymbol{w}^i)$. Such states typically suggest the parameters of system are at fixed point in the iteration of renormalization, meaning the system's parameters don't change with scale. Fig. {\ref{fig:framework}} show the schematic diagram of the whole framework.

Extending to the network in non-Euclidene space, $S$ is regarded as an S-order neighbors, $\boldsymbol{w}^i$ describes the mechanism of interaction among the subject $i$ and these neighbors, and $P^i$ is the mapping operation of the S-order neighbors in the T time range to the macroscopic state.


An effective verification method for self-similar dynamics can be carried out using the following steps.

1. Coarse-grain the microstate $\boldsymbol{\Tilde{X}}^i_t$ to obtain $\boldsymbol{y^i_t}$. Then evolve $\boldsymbol{y^i_t}$ by one step to get $\boldsymbol{y^i_{t+1}}$

\begin{equation}
\label{eq:self_similarity1}
    \boldsymbol{y^i_{t+1}} = \boldsymbol{y^i_{t}} + f^i[P^i(\boldsymbol{\Tilde{X}}^i_t)]
\end{equation}

2. Coarse-grain the microstate $\boldsymbol{\Tilde{X}}^{i}_{t+1}$ directly
\begin{equation}
\label{eq:self_similarity2}
    (\boldsymbol{y^i_{t+1}})' = P^i [\boldsymbol{\Tilde{X}}^{i}_{t+1}]
\end{equation}

Compare $ \boldsymbol{y^i_{t+1}} $ from step 1 and $(\boldsymbol{y^i_{t+1}})'$ from step 2. If consistent results are obtained, the dynamics can be considered self-similar. For general dynamics, we can get the difference between the two. In other words, the discrepancy between $ \boldsymbol{y^i_{t+1}} $ and $(\boldsymbol{y^i_{t+1}})'$ serves as a metric for the extent of self-similarity for a dynamics. This is referred to as the consistency in subsequent discussions. In the experiments, we use the mean square error (MSE) as a measure of consistency, shown as Eq. \ref{eq:consistency}.
\begin{equation}
\label{eq:consistency}
    consistency = MSE(\boldsymbol{y^i_{t+1}}, (\boldsymbol{y^i_{t+1}})')
\end{equation}

It is noteworthy that the premise for the consistency metric to be meaningful is that the coarse-graining strategy $P$ itself is meaningful, i.e., the macro states we obtain are not trivial.

\subsection{Framework}
\label{se:framework}
According to the definition, we utilize a multiscale neural network approach, comprising two main components: a dynamics learner and a coarse-graining learner. 

\textbf{Dynamics Learner}: Using micro-level data $\boldsymbol{X}$ to learn the dynamic rules $f(\boldsymbol{\theta_1})$, The neural network parameters effectively serve as the dynamic parameters $\boldsymbol{w}_{Nei}$. The optimization goal is to minimize the difference between the actual value $\boldsymbol{x}_{t+1}$  and the predicted value $\hat{\boldsymbol{x}}_{t+1}$. Considering the differences in data dimensions between the micro and macro scales—with macro variables typically being fewer—it is important to design neural networks that demonstrate spatio-temporal translational invariance. Only in this manner can we potentially apply the dynamics learner, trained on micro-level data, to macro-level data as well.

\textbf{Coarse-Graining Learner}: Learning the mapping from the microstates 
$\boldsymbol{X}$ to the macrostates $\boldsymbol{Y}$, expressed as $\boldsymbol{Y}=P_{Nei}(\boldsymbol{X},\boldsymbol{\theta_2})$. The objective is to ensure that the learned macrostates $\boldsymbol{Y}$ align with the dynamics $f(\boldsymbol{\theta_1})$, that is, the form and the parameters of macro-dynamics should be same as the micro-dynamics, and we could still get an accurate prediction from $\boldsymbol{y}_t$ to $\boldsymbol{y}_{t+1}$ according to $f(\boldsymbol{\theta_1})$. In other words, the self-similarity in system dynamics is incorporated as a constraint in the model design, guiding the Coarse-graining learner to adhere to this constraint.

\begin{figure*}
    \centering
    \includegraphics[width=1\linewidth]{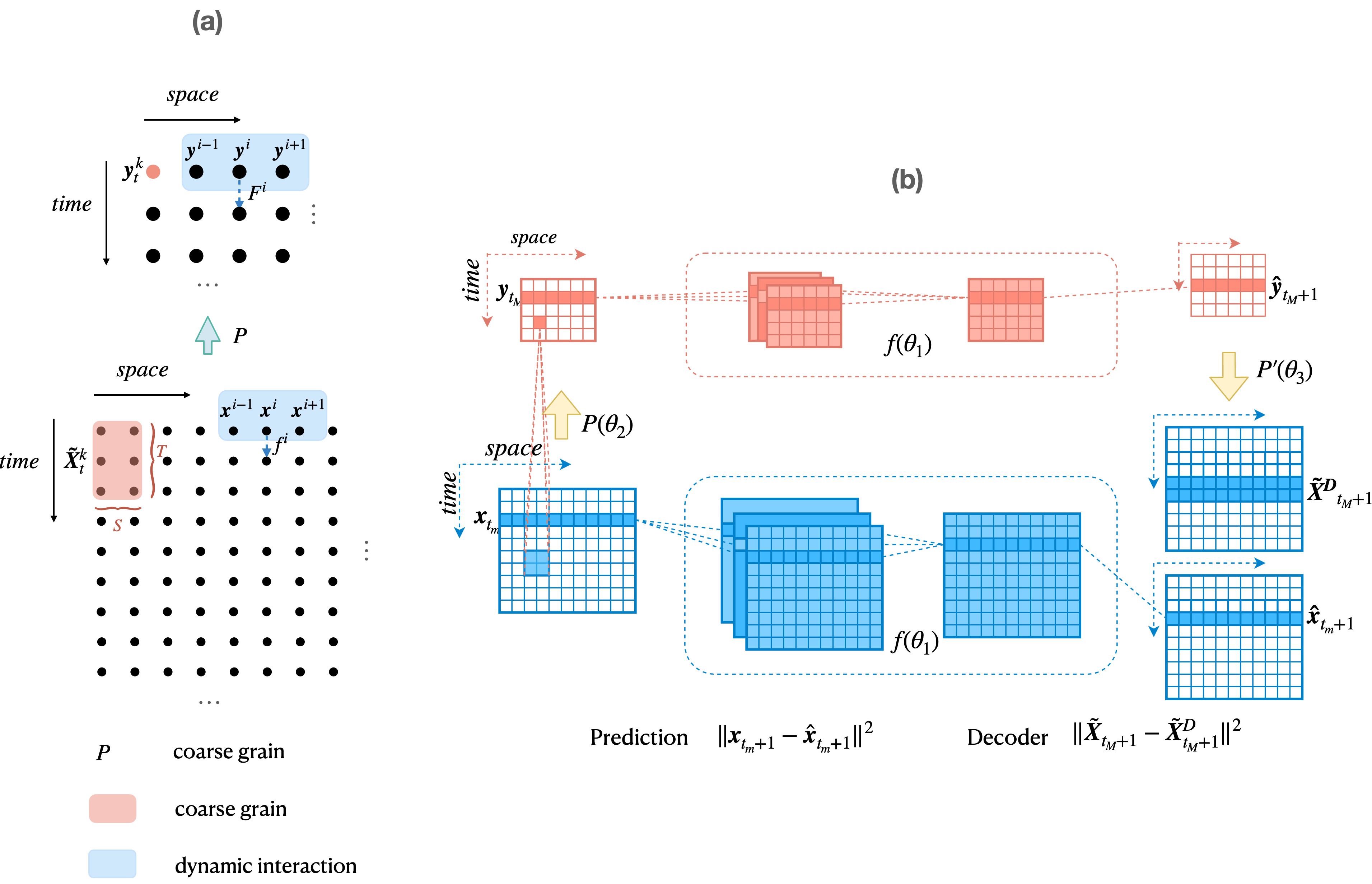}
    \caption{(a).\textbf{Diagram of for dynamics and coarse-graining in 2D lattice systems}. The blue area represents the range of dynamical interactions, here is the nearest neighborhood interaction as an example. Blue dashed arrow means $\boldsymbol{x}^i$ or $\boldsymbol{y}^i$ evolves one step with dynamic $f^i$ or $F^i$. The red area represents a basic unit of microstate $\boldsymbol{\Tilde{X}}^{k}_{t}$ and it will be coarse-grained as a macrostate $\boldsymbol{y}^k_t$, displaying as a red point in the figure above. Both dynamics and coarse-graining operator could apply to any of other states and areas, which means the operators are homogeneous. (b).\textbf{CNN model applied to our framework}. We have two level neural network framework. The first level aims to capture the microscopic dynamics $f$ using the microstate which are presented blue in the bottom of the figure, while the second one represents the macroscopic state and dynamics $F$, which are red on the top. Both dynamics receive time-series data from their respective levels as input and produce predictions for future outcomes. As mentioned to in the main text, both the micro and macro dynamics share the same neural network  structure and parameters, as depicted by $f(\theta_1)$ in the figure. Given the discrepancy in data dimensions between the micro and macro levels—typically with macro variables being fewer — it necessitates the construction of neural networks that exhibit spatio-temporal translational invariance.  A coarse-graining learner $P(\theta_2)$,connects the two, facilitating the transformation from microscopic to macroscopic data. To prevent macrostates from falling into trivial state in the training process, we added a decoder $P(\theta_3)$ after the macroscopic prediction. This decoder constrains its output to closely resemble the microstates. After training, the decoder can be deactivated. }
    \label{fig:framework_cnn}
\end{figure*}

The model framework is illustrated in Fig \ref{fig:framework}. We aim for the predicted microstate $\hat{\boldsymbol{x}}_{t_m+1}$ to closely approximate the actual state $\boldsymbol{x}_{t_m+1}$, this is the first loss function of our framework in order to optimize the dynamical Learner (Eq.\ref{eq:loss1}). For the coarse-grained learner, we incorporate a third decoder neural module $P'(\boldsymbol{\theta_3})$ to prevent learning trivial coarse-grained rules, such as mapping all states to zero. This is done to ensure that the macro-states $\boldsymbol{y}_{t_M+1}$can be effectively decoded back into microstates $\boldsymbol{\Tilde{X}}_{t_M+1}$. Therefore, the second learning objective of our framework is to make the reconstructed state $\boldsymbol{\Tilde{X}}^{D}_{t_M+1}$ as close as possible to the actual state $\boldsymbol{\Tilde{X}}_{t_M+1}$ (Eq.\ref{eq:loss2}). Fig \ref{fig:framework_cnn} (a) visualizes the whole process with spatial dimension $d=1$. It can be easily extended to higher dimensions.
%
\begin{equation}
    \label{eq:loss1}
    \boldsymbol{\theta_1}^* = arg min(\| {\boldsymbol{x}}_{t_m+1}-f(\boldsymbol{x}_{t_m},\boldsymbol{\theta_1}) \|^2 )
\end{equation}

\begin{eqnarray}
    \label{eq:loss2}
    \boldsymbol{\theta_2}^*,\boldsymbol{\theta_3}^*  &=& arg min( \| \boldsymbol{\Tilde{X}}_{t_M+1}-\boldsymbol{\Tilde{X}}^{D}_{t_M+1} \|^2) \\
    &=& arg min(\| \boldsymbol{\Tilde{X}}_{t_M+1}-P'(\boldsymbol{\hat{y}}_{t_M+1},\boldsymbol{\theta_3}) \|^2) \nonumber \\
    &=&  arg min(\| \boldsymbol{\Tilde{X}}_{t_M+1}-P'(f(\boldsymbol{y}_{t_M},\boldsymbol{\theta_1}^*),\boldsymbol{\theta_3}) \|^2) \nonumber \\
    &=& arg min(\| \boldsymbol{\Tilde{X}}_{t_M+1}-P'(f(P(\boldsymbol{\Tilde{X}}_{t_M},\boldsymbol{\theta_2}),\boldsymbol{\theta_1}^*),\boldsymbol{\theta_3}) \|^2) \nonumber
\end{eqnarray}

Our multiscale framework is inspired by the design in \cite{zhang2022neural}, while they used Invertible Neural Networks (INN) in their encoder and decoder to ensure favorable mathematical properties. In our approach, the encoder and decoder are two distinct neural networks. We made this choice to circumvent the efficiency issues encountered during the training of INN. Additionally, our model training is divided into two stages. The first stage achieves optimal dynamics, which is then applied at the macro level in the second stage, rather than training them concurrently. We adopted this approach to ensure the functions of the dynamics and the encoder remain independent and don't become intertwined.

\section{Experiments}
\label{se:experiments}
In the following part, we will validate the effectiveness of our frame using two examples. Cellular automata serve as a representation of deterministic dynamic systems, while diffusion dynamics stand for stochastic dynamic systems. 

\subsection{1d Cellular Automata}
\label{se:1dCA}
In 1D cellular automata, there are 256 different nearest-neighbor rules, which could create lots of amazing different patterns. In literature \cite{Israeli2006}, the authors have summarized there are 21 rules can exhibit self-similarity through appropriate coarse-graining techniques. The coarse-graining methods in the article were found through manual search. The dynamics and evolving patterns of cellular automata correspond, so according to Eq.\ref{eq:self_similarity1} and Eq.\ref{eq:self_similarity2}, a system is considered self-similar if the macroscopic patterns after coarse-graining resemble the microscopic patterns. We use the classification in this literature as the ground truth to validate the effectiveness of our framework in deterministic dynamic systems.

\textbf{Dynamics Learner}: We configure the dynamical learner as a 1D convolutional neural network with a kernel size of 3. Given that cellular automata operate based on nearest-neighbor interactions, a single convolutional layer is sufficient to capture the complete dynamics. We employ two kernels and combine them through a linear mapping to generate predictions for the next time step, just as Fig.\ref{fig:framework_cnn}(b) has shown.

\textbf{Coarse-Graining Learner}: For the coarse-graining learner, we utilize a 2D convolutional neural network and kernel sizes of 2x2, 3x3, and 4x4 for all rules. The choice of square-shaped kernels is motivated by the need for consistency in both spatial and temporal scales after each coarse-graining step. In nearest-neighbor cellular automata, if we want the dynamics to continue to be based on nearest-neighbor interactions after coarse graining, the temporal and spatial scales should be equal, that is $S=T$. Because of the same value between $S$ and $T$, we rename them as group size $N$ for convenience in the following part.

\begin{figure}
    \centering
    \includegraphics[width=1\linewidth]{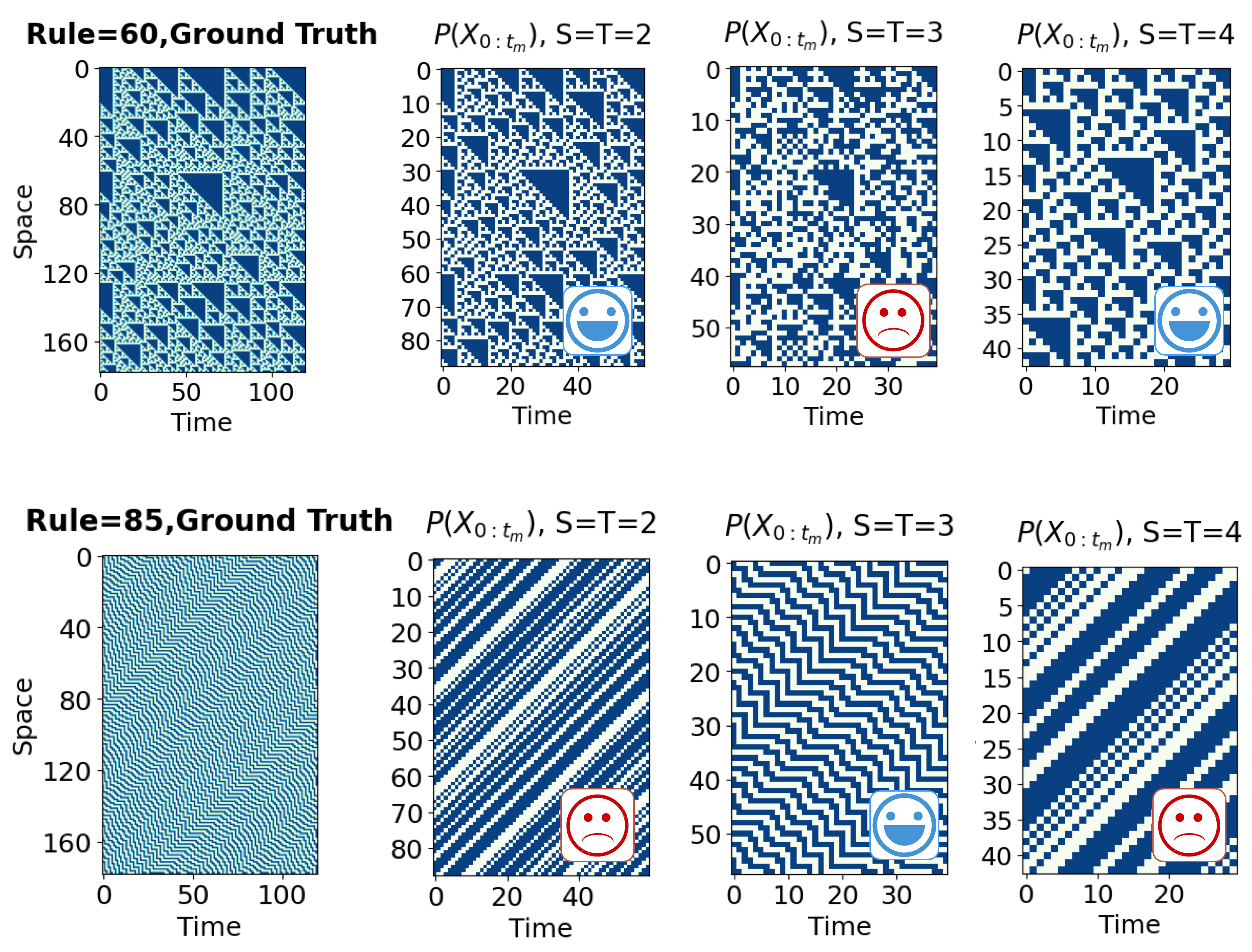}
    \caption{\textbf{Examples of CA dynamic renormalization based on self-similar constraints} (rule = 60 and rule = 85). The first row is the result of rule 60. They are ground truth and renormalization results of three different scales with $S = T = 2,3,4$, respectively. The second row is the result of rule 85, and also are ground truth and renormalization results of three different scales with $S = T = 2,3,4$. We could see except $S = T = 3$, rule 60 can be renormalized as same rule in macro level, and rule 85 can be renormalized a self-similar dynamics only for $S = T = 3$.}
    
    \label{fig:ca_ex}
\end{figure}

Rule 60 and 85 are taken as examples to show the learning result of renormalization. In the case of specifying group size, our framework can identify whether the current rule is self-similar dynamics according to the pattern (Fig.\ref{fig:ca_ex}). We can see that for rule 60, except group size N equals to 3, both N=2 and N=4 can map the rule to a smaller scale and keep the dynamics unchanged. And rule 85 could be renormalized as a self-dynamics only when N=3. Table 1 shows more complete results, which shows our framework can identify all the self-similar 21 rules,  find the proper group size and learn coarse-grain mapping. At this time, The consistency error is almost equal to 0, because the dynamics of CA can be completely captured for the neural network. Experiments also show that we can learn the dynamics of all 256 CA rules with nearly 100 \% accuracy. At this time, the consistency can directly give the identification of yes or no for self-similarity. As for non-self-similar rules, they are not fully presented here, but experiments show that they do fail self-similarity verification according to Eq.\ref{eq:consistency}


\begin{table}
    \begin{center}
        \label{tab:ca}
        \caption{\textbf{Consistency for different CA rules}}. Here we show applying our framework to cellular automata with all 21 self-similar rules and identifying what is the proper group size N if we want to get self similarity dynamics. The number in the table is the MSE of $ \boldsymbol{y^i_{t+1}} $ and $ (\boldsymbol{y^i_{t+1}})' $ according to Eq.\ref{eq:consistency}.  \textbf{0} means the dynamics are self-similar.
        \begin{ruledtabular}
        \begin{tabular}{c|c c c |c| c c c} 
             \textbf{Rule} & N=2& N=3& N=4 &\textbf{Rule} & N=2& N=3& N=4\\
            \hline
             0&\textbf{0.0}&\textbf{0.0}&\textbf{0.0}&
             170&\textbf{0.0}&\textbf{0.0}&\textbf{0.0} \\
             \hline
             15&0.5&\textbf{0.0}&0.5116  &
            165&\textbf{0.0}&0.4911&\textbf{0.0} \\
             \hline
            51&0.5&\textbf{0.0}& 0.5116  &
            192&\textbf{0.0}&\textbf{0.0}&\textbf{0.0}  \\
             \hline
            60&\textbf{0.0}&0.4786&\textbf{0.0} &
            195&\textbf{0.0}&0.4520&\textbf{0.0} \\
             \hline
             85& 0.5 &\textbf{0.0}& 0.4884&
            204&\textbf{0.0}&\textbf{0.0}&\textbf{0.0} \\
             \hline
              90&\textbf{0.0}&0.5246&\textbf{0.0}&
            238&\textbf{0.0}&\textbf{0.0}&\textbf{0.0}\\
             \hline
              102&\textbf{0.0}&0.2990&\textbf{0.0}&
            240&\textbf{0.0}&\textbf{0.0}&\textbf{0.0} \\
             \hline
              128&\textbf{0.0}&\textbf{0.0}&\textbf{0.0}&
            252&\textbf{0.0}&\textbf{0.0}&\textbf{0.0} \\
            \hline
            136&\textbf{0.0}&\textbf{0.0}&\textbf{0.0}&
            254&\textbf{0.0}&\textbf{0.0}&\textbf{0.0} \\
            \hline
            150&\textbf{0.0}&0.7723&\textbf{0.0}&
            255&\textbf{0.0}&\textbf{0.0}&\textbf{0.0}\\
            \hline
            153&\textbf{0.0}&0.5407&\textbf{0.0}&-&&& \\
        \end{tabular}
        \end{ruledtabular}
    \end{center}
\end{table}

\subsection{Reaction-Diffusion Process}
\label{se:diffusion}
The reaction-diffusion equation is used to describe the processes of material diffusion and reaction. Typically, this equation can be used to study the distribution changes of materials over time and space. Generally speaking, the standard form of the reaction-diffusion equation is Eq.\ref{eq:diffusion_process}.
\begin{equation}
    \label{eq:diffusion_process}
    \frac{\partial C}{\partial t} = D \frac{\partial^2 C}{\partial x^2} + R(C)
\end{equation}


$C(x,t)$ represents the probability distribution of particle concentration $C$ at location $x$ at time $t$. $D$ is the diffusion coefficient, and $R(C)$ is a reaction rate function related to concentration $C$, which describes the reaction process of a substance. Let us think about the simpler case, that is $R(c) = 0$, which means the total concentration is conserved throughout the diffusion process. In that case,  when appropriately coarse-grained, the diffusion coefficient $D$ of the diffusion process remains unchanged, indicating that the system have a self-similar dynamics.  We can use this system to test whether our framework can be applied to stochastic dynamics.
%


\textbf{Dataset}: In numerical simulations of the diffusion process, it is intuitive to assume that the smaller the time interval $\Delta t$ and the spatial interval $\Delta x$, the higher accuracy of the simulation. As the intervals approach zero($\Delta t$ and $\Delta x \to 0$), the diffusion dynamics converge towards perfect self-similarty. We've generated data using different time-space intervals: $\Delta t=1,0.1,0.04,0.01$ and for $\Delta t \to 0$--derived from the analytical solution. All simulations evolved from $t=0$ to $t = 100$ within a spatially bounded system of size 100, with periodic boundary conditions. Fig.S1 (a) provides a visualization of the data. 

\textbf{Dynamics Learner}: We modeled the dynamics using a 1D convolutional neural network (CNN). Without any prior knowledge, we experimented with different kernel sizes to assess their impact on prediction accuracy. The optimal accuracy was achieved with a kernel size of 5 (as shown in Fig.\ref{fig:diffusion_mae} (a) left). A kernel size of 3 and 7, on the other hand, offered a good margin of error tolerance when $\Delta t != 0$ (Fig.\ref{fig:diffusion_mae} (a) right ). In the absence of specific instructions of this paper, we standardized the kernel size to 7. Fig.S1 (b) depicts the state transition vectors learned by the dynamics learner for simulation data of varying time-space interval. It's evident that the higher the accuracy of the simulation data, the closer the dynamics transition vector is to a Gaussian distribution. 

\textbf{Coarse-Graining Learner}: For a simple diffusion process, the relationship between particle movement in time and space satisfies $x \propto \sqrt{t}$. This means that, when renormalizing the diffusion process, the space-time scaling relationship needs to be $S \propto \sqrt{T}$  to potentially yield the same dynamics. Fig S3 shows the results if the relationship is not $x \propto \sqrt{t}$, Neither the accuracy of the predictions nor the scaling of different scales can be reproduced. We adopted a spatio-temporal convolutional approach for learning. Specifically, we utilized two separate 1D convolutional neural networks for time and space dimensions. By adjusting the kernel sizes for convolution and pooling, we can coarsely grain the data into a more compact space.

Firstly, we wanted to validate whether our framework could function effectively in the diffusion process. Applying the aforementioned learner to the data where $\Delta t \to 0$, Fig.\ref{fig:diffusion_visualization}(a) displays the visualization comparison between the ground truth and prediction result, and Fig.\ref{fig:diffusion_visualization}(b)  shows the visualization of consistency for dynamics between $\boldsymbol{y}^i_{t+1}$ and $(\boldsymbol{y}^i_{t+1})'$.  More quantitatively, we found that the prediction error for the dynamics could be extremely low, with an MSE approaching the magnitude of $10^{-6}$ (Fig.\ref{fig:diffusion_mae} (a)). {Fig.\ref{fig:diffusion_mae} (c) displays the dynamic consistency for different $\Delta t$ and we found when $\Delta t \to 0$, the error between $\boldsymbol{y}^{i}_{t+1}$  and $(\boldsymbol{y}^{i}_{t+1})'$  is lowest compare to other situation and being only near $ 10 ^{-4}$ .

We also calculate the relationship between the time $t$ and the diffusion mean square distance in learned macroscopic state, and the slope of the relationship is related to the diffusion coefficient $D$. Our results show that the macroscopic and microscopic diffusion coefficients are very consistent (Fig.\ref{fig:diffusion_visualization} (b)). This phenomenon suggests that we not only can judge which dynamics are closer to being self-similar based on the consistency, but also can learn the correct coarse-graining strategy for this self-similar dynamics and extract the power law exponent

\begin{figure}
    \centering
    \includegraphics[width=1\linewidth]{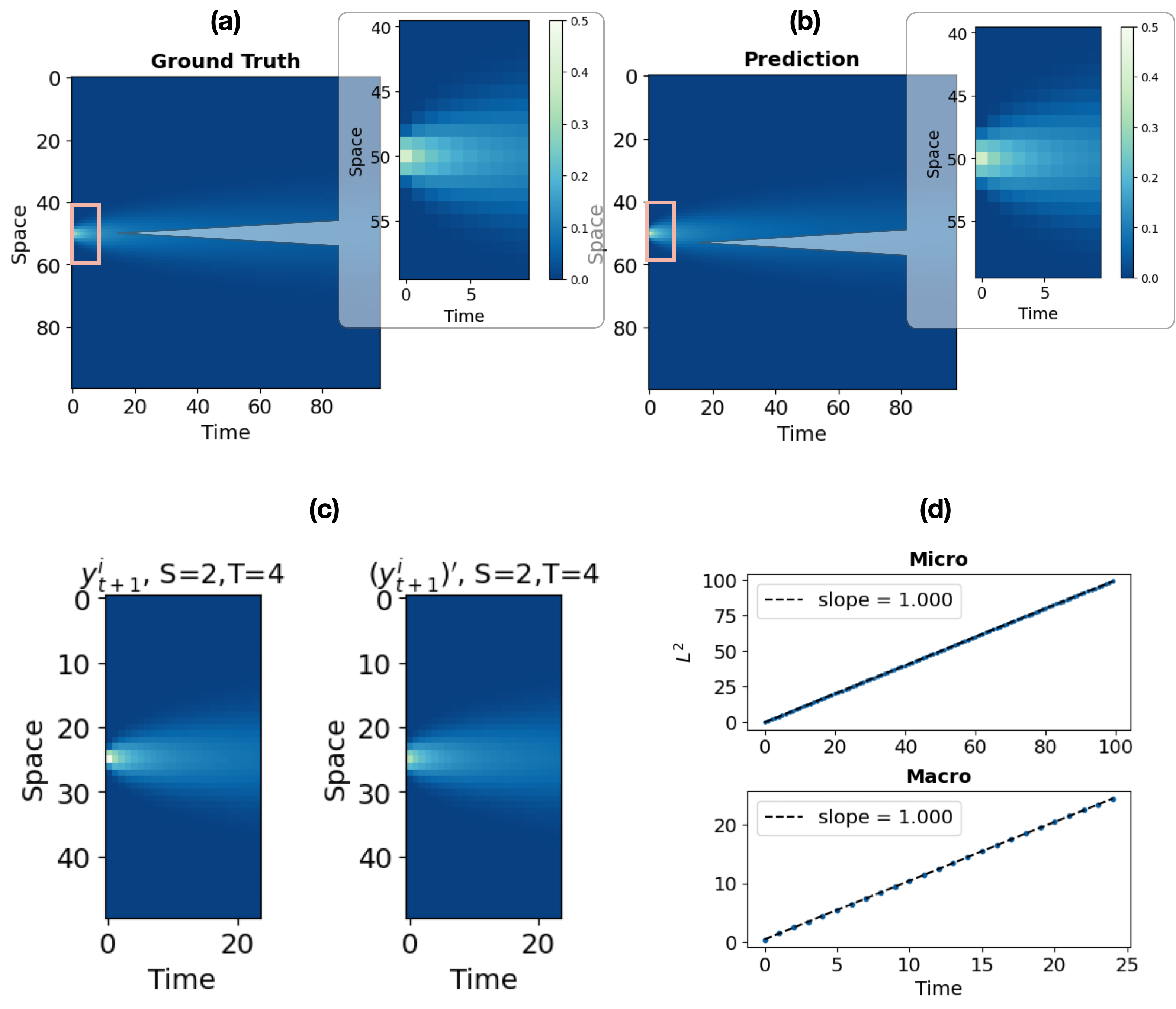}
    \caption{\textbf{Visualization of dynamic prediction and consistency results for diffusion process.} (a) shows the ground truth and prediction of microstate when $\Delta to \to 0$, respectively. Subplots are the enlarging version of the red square in order to get more clear visualization. (b) are visualization of consistency for dynamics \iffalse the verification of self-similarity\fi when $\Delta t_0 \to 0$, same as CA before. We set $S=2$ and $T=4$ as example, while  $S=3$ and $T=9$ also have the similar results, which we don't show here due to space constraints. (c) are the verification of the scaling relationship between time $t$ and average diffusion length $L$ in microstates and macrostates respectively.}
    \label{fig:diffusion_visualization}
\end{figure}


Further, by using different $\Delta t$ values as training data, we can derive various models and observe how the model outcomes differ. Fig.\ref{fig:diffusion_mae} (a) illustrates the changes in the convolutional kernels learned by the dynamics as $\Delta t$ varies. We noticed that the kernels gradually converge to a bell-shaped curve. Simultaneously, the error in the dynamics decreases as $\Delta t$ diminishes, as shown in Fig.\ref{fig:diffusion_mae} (a). Similarly, the error from reconstruction also lessens as $\Delta t$ gets smaller (Fig.\ref{fig:diffusion_mae} (b)), with a notably pronounced advantage when learning on self-similar dynamics. Note that all experiments here are the average results of three experiments, and the error bar is not obvious in the figure due to its small value.

\begin{figure}
    \centering
    \includegraphics[width=1\linewidth]{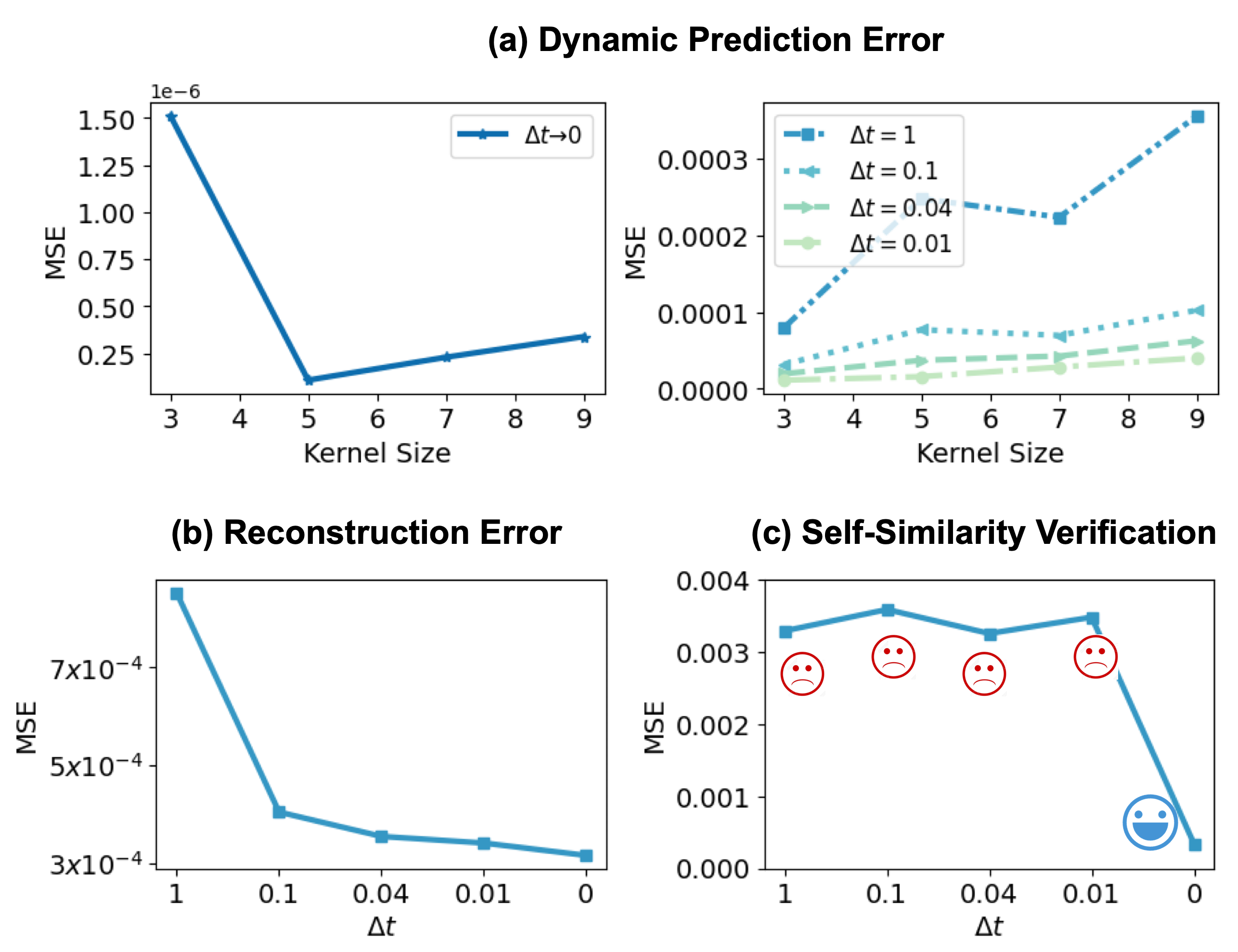}
    \caption{\textbf{Comparison of different $\Delta t$ and quantitative results of dynamic consistency}. (a) Testing the influence of different kernel sizes on dynamic accuracy. we found for the self-similar dynamic (left), kernel size = 5 and 7 could have the relative great results, while kernel size = 3 and 7 are better for none self-similar dynamic (right). So in the absence of any specific instructions, our experimental setup is kernel size = 7. (b). MSE for decoder module and the reconstruction error of self-similar dynamics is minimal. (c). Quantitative self-similar verification. We could found our framework is very effective for identifying self-similar dynamics. Each curve is the average result of three experiments, and the error bar is not obvious in the figure due to its small value}
    \label{fig:diffusion_mae}
\end{figure}

\subsection{Vicsek Model}}
A more convincing validation of our framework is to apply it to critical dynamical systems. The Vicsek model is a simple yet influential computational model used to study the emergent behavior of self-propelled particles, which is often used to represent the collective motion observed in biological systems, such as flocks of birds, schools of fish, and swarms of insects. Developed by Vicsek\cite{Vicsek1995}, the model illustrates how complex patterns of movement can emerge from simple rules followed by individuals, without any central coordination. The Vicsek model is characterized by two control parameters: the density of the system and the level of noise. Particles move at a constant speed and interact with their nearby neighbors using simple rules, continuously updating their positions within the system. The classic Vicsek model is mathematically represented as follows:
\begin{equation}
    \label{eq:vicsek1}
    \boldsymbol{x}_i(t+\Delta t) = \boldsymbol{x}_i(t) + \boldsymbol{v}_i(t) \Delta t 
\end{equation}

\begin{equation}
    \label{eq:vicsek2}
    \theta_i(t+\Delta t) = <\theta_i(t)>_r + \eta \xi_i
\end{equation}

where $\boldsymbol{x_i}(t)$, $\boldsymbol{v_i}(t)$ and $\theta_i(t)$ is the position, velocity and angle of particle $i$ at time $t$, respectively. $<\theta_i(t)>_r$ is the neighborhood's average direction of particle $i$ within radius $r$, $\eta$ is the noise intensity, $\xi_i$ is a random variable sample from uniform distribution. The average orientation of the particles $\phi = \frac{1}{N}\sum^N_i \theta_i$ changes with variations in the system's noise and density. Within certain parameter regions, phase transitions and critical phenomena can occur, with the order parameter being the system's average velocity. We employ the Vicsek model as a new example to validate our framework, as the first step for exploring critical dynamic systems.

\begin{figure}
    \centering
    \includegraphics[width=1\linewidth]{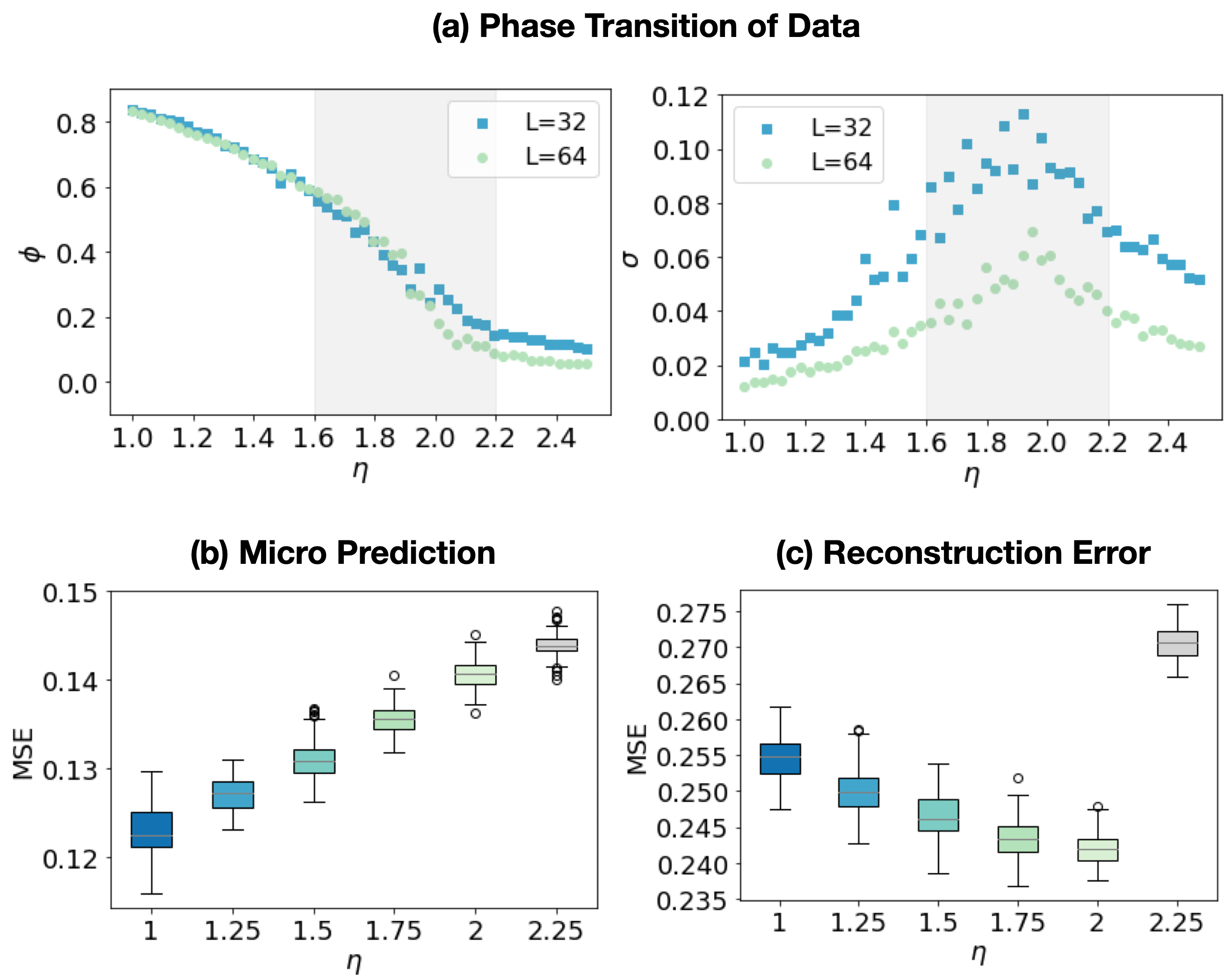}
    \caption{\textbf{Results of vicsek model}. (a) is the order parameter $\phi$ (left) and its standard deviation (right) change with noise intensity $\eta$ in $L = 32$ and $64$. we can roughly estimate from the figure that the critical region to be in the range of $\eta=1.6$ to $2.2$. (b) is the MSE for five different dynamic learners. (c) is the decoder MSE for coarse-graining learners. Both (b) and (c) were repeated experiments for 3 times, and a total of 3000 snapshoot data were used to draw the box diagram.}
    \label{fig:vicsek_results}
\end{figure}

\textbf{Dataset}: In our simulations, we set the model parameters such that the absolute velocity is 1, and each particle's field of vision radius is 1, with the particles' angles initialized as random. Particle positions are updated according to Eq. \ref{eq:vicsek1} and Eq.\ref{eq:vicsek2}. The orientation of each particle update is sampled in $[-\pi/2,\pi/2]$. We tested the outcomes under different noise conditions for systems of the same density ($\rho=0.3$) but different sizes ($L=32$ and $L=64$). The noise intensity was chosen to be $\eta=1,1.25,1.5,1.75,2,2.25$. Each noise can be used to train a new model. From Fig \ref{fig:vicsek_results}(a), it can be seen that the critical region is about between $\eta=1.6$ to 2.2. It is noteworthy that the Vicsek model's mechanism involves interactions with surrounding neighbors, which change at each moment. To simplify the problem, we latticeized the multi-agent model, which does not affect the calculation of the order parameter at micro level. The specifics of the data processing details are shown in the Appendix 1. Ultimately, the data used for learning is a time series of size $L \times L$, with three feature dimensions: the density , and the velocity components in two directions at each site at time $t$. Fig S2 shows the visualization of our data.

\textbf{Dynamics Learner}: We utilized a 2D CNN for modeling the dynamics. Given that the field of vision radius was set to 1 during data generation, we configured the kernel size of the dynamics learner to be $2 \times 2$.

\textbf{Coarse-Graining Learner}: For the coarse-graining learner, we employed 3D convolution to simultaneously process time and space. We experimented with $T=S=2$ and $T=4,S=2$, corresponding to convolutional kernel sizes of $2 \times 2 \times 2$ and $4 \times 2 \times 2$, respectively. According to the work \cite{Cavagna2017}, the dynamical critical exponent $z$ of Vicsek model $\approx 2$. which means the latter kernel could model better than the first one.

In our preliminary exploration, we found that the reconstruction error of the model is indeed related to the system's criticality. For the micro prediction error, as evident from Fig. \ref{fig:vicsek_results}(b), there is a significant positive correlation with the values of noise, which is because the higher the noise, the weaker the predictability of the system, which is aligns with our experience. As for the reconstruction error, the model exhibits the lowest error in the critical region. In both the ordered and disordered states, the reconstruction error significantly increases, shown as Fig.\ref{fig:vicsek_results}(c). When the noise intensity $\eta > 2.25$, the model is already not convergent because the data is too disordered, so we do not plot the figure with $\eta > 2.25$. Both Fig.\ref{fig:vicsek_results}(b) and (c) were repeated experiments for 3 times, and a total of 3000 snapshoots were used to draw the box diagram. From the figure, we can conclude that our model can indeed identify critical regions. In the supplementary materials, we also demonstrate that in scenarios where $T\neq S^2$ , meaningful patterns cannot be discerned from the experimental outcomes.

Fig.S4 showcases the results of dynamic consistency. We observed that this metric reaches its lowest at $\eta=2.25$. As previously mentioned, the effectiveness of the consistency metric is predicated on the macro variables not being trivial. In fact, at the disordered state for vicsek model, the macro state has already entered trivial state, rendering the consistency metric meaningless.



\subsection{Self-similarity vs. non-self-similarity}

\begin{figure}
    \centering
    \includegraphics[width=1\linewidth]{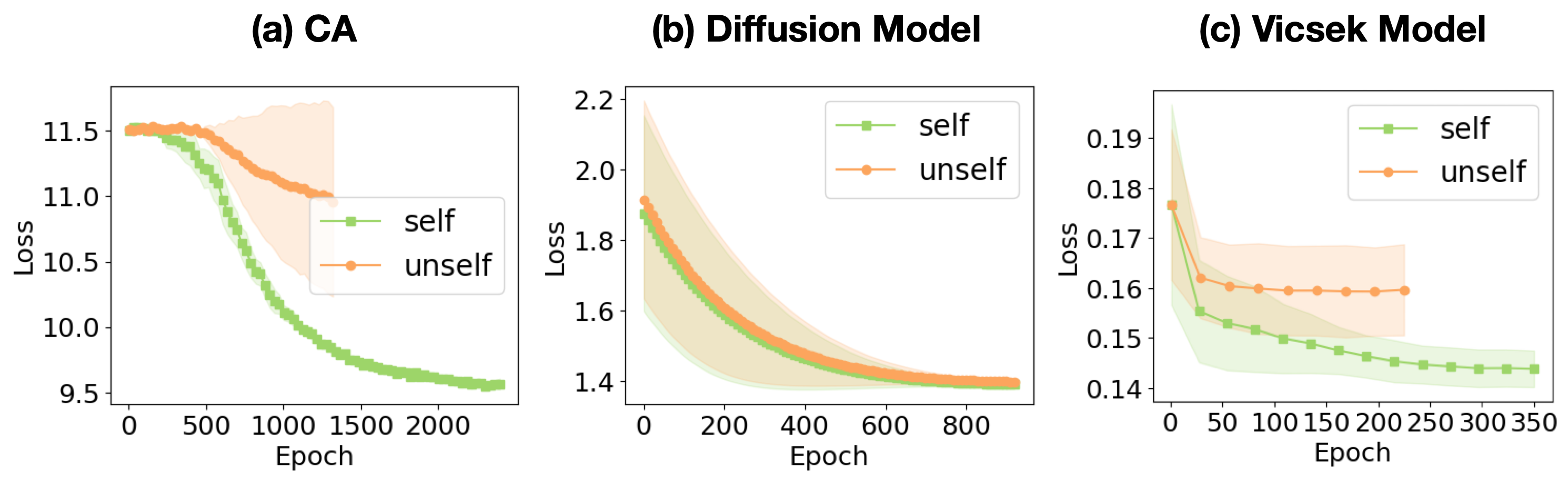}
    \caption{\textbf{Comparison of loss functions for self-similar and non-self-similar frameworks in all example models.} Blue lines are the loss of self-similarity while the green lines for the none-self-similarity. All the lines here are the average of the 5 experiments, and the shaded part represents the standard deviation. Because the epoch of convergence of each experimental model is different, and the shortest epoch is taken when drawing this graph, it seems that the curve has not converged yet. But we can ensure that the loss will not decrease again in each experiment before stopping. }
    \label{fig:self_vs_unself}
\end{figure}

In this section, we compare the experimental results of our self-similar framework with those of a non-self-similar framework under the same parameters. A non-self-similar framework means that the macroscopic dynamics do not share parameters with the microscopic ones, and the parameters for the macroscopic dynamics are trained simultaneously with the coarse-graining strategy learning process. Fig.\ref{fig:self_vs_unself} shows the decrease in validation set loss in experiments with three model. The images display the average results of five experiments. In all types of experiments, the self-similar framework outperforms the non-self-similar framework.

Our model fails to converge most of the time when setting as non-self-similarity, indicating that the learned macroscopic dynamics and coarse-graining strategy are meaningless. In fact, it is the inclusion of the self-similarity prior that allows us to learn the correct coarse-graining strategy. Fig. S5 shows the test results after training with a non-self-similar framework using CA model Rule 85 as an example (which belongs to self-similar dynamics with a group size = 3). We find that the coarse-graining maps the microscopic states into a trivial state, that is, for any $\boldsymbol{X}$, $P(\boldsymbol{X})=0$ or $1$. Thus, the macroscopic dynamics can only learn a meaningless mapping from $0$ to $0$ or $1$ to $1$. This is precisely why the model does not converge.


\section{Conclusion and Discussion}

Analyzing complex systems from a multi-scale perspective is what sets them apart from other systems. We often find that fascinating phenomena like emergence are linked to the scale of the system. The motivation behind constructing self-similar dynamics is to find a scale-independent kernel for a dynamics system. Although our exploration is still in its early stages, our model holds significant potential.

In summary, we introduce a multiscale neural network framework that integrates self-similarity priors for modeling complex dynamic systems. We can identify if the dynamic systems are self-similar, whether deterministic or stochastic dynamics. And our framework can automatically capture scale-invariant kernel in dynamics based on the setting of homogeneity, which means we can use fewer parameters to model the dynamic system. In our experiments, there are only a dozen of parameters in our neural model, but it also can capture the precise dynamics.

In fact, we think our model also has a strong relationship with renormalization group. Renormalization group signifies the alteration of models describing systems as scales vary, with dynamical renormalization focusing on the dynamical models. While physicists have historically achieved noteworthy successes with renormalization in statistical models, recent years have seen substantial progress in marrying machine learning with renormalization frameworks. However, research on dynamical renormalization has often been overly manual, necessitating the search for new appropriate renormalization strategies when faced with a novel system. Our work offers a potential framework for the fusion of dynamical renormalization with machine learning. The incorporation of self-similar prior knowledge suggests we are modeling fixed points or the vicinity of the fixed point in the dynamical parameter space. We firmly believe that data-driven multi-scale modeling approaches will present viable solutions for automating dynamical renormalization, paving a significant avenue for the integration of machine learning and statistical physics.

In the discussions of this paper, we have used what is referred to as dynamic consistency to judge self-similarity in most of our cases. In fact, recent theories on causal emergence have proposed a metric for dynamic causality—the Effective Information (EI)\cite{Hoel2013}. This metric, from an information theory perspective, comprehensively measures two dimensions of a dynamic system: its determinacy and degeneracy. When EI is used to assess dynamics across different scales, it can serve to determine the occurrence of emergence. Recent studies have also explored the idea of using EI as an optimization objective to find the so-called optimal scale\cite{zhang2022neural,yang2023finding}. Theoretically, EI can also serve as an indicator of dynamic self-similarity. The EI of self-similar dynamics should be consistent across different scales. This could act as a form of cross-validation for our framework. These discussions are worth further exploration in our future work.

\section*{Acknowledgements}
We would like to express our gratitude to Swarma Club for their support towards our paper, as well as the constructive discussions contributed by Zhangzhang, Mingzhe Yang, Zhongpu Qiu, Mingze Qi. During the preparation of this work the author(s) used GPT-4 in the process of writing paper in order to make the English expression more accuracy and fluency. After using this tool/service, the author(s) reviewed and edited the content as needed and take(s) full responsibility for the content of the publication.

\section{AUTHOR DECLARATIONS}
The authors have no conflicts to disclose
\section{DATA AVAILABILITY}
The data that support the findings of this study are available from the corresponding author upon reasonable request

\section{References}
\bibliography{aipsamp}

\end{document}